\documentclass{article}

\usepackage[T1]{fontenc}
\usepackage[utf8]{inputenc}
\usepackage[english]{babel}

\usepackage[letterpaper,top=2cm,bottom=2cm,left=3cm,right=3cm,marginparwidth=1.75cm]{geometry}

\usepackage{microtype}

\usepackage{amsmath}
\usepackage{booktabs}
\usepackage{graphicx}
\usepackage{caption}
\usepackage{subcaption}

\usepackage{listings}
\lstdefinestyle{query}{
  basicstyle=\ttfamily\footnotesize,
  breaklines=true,
  breakatwhitespace=false,  
  columns=fullflexible,     
  keepspaces=true,
  showstringspaces=false,
  frame=none,
  xleftmargin=0pt, xrightmargin=0pt
}

\usepackage[hyphens]{url}
\usepackage[colorlinks=true,allcolors=blue]{hyperref}

\usepackage[numbers,sort&compress]{natbib} 
\title{Designing and Analysing Argument Mining Pipelines: \\Towards a Comprehensive Assessment}
\author{Siddharth Bhargava \\
Fondazione Bruno Kessler, Trento, Italy \\
Universidade da Coruña, A Coruña, Spain \\
\texttt{sbhargava@fbk.eu}
\and
Sara Tonelli \\
Fondazione Bruno Kessler, Trento, Italy \\
\texttt{satonelli@fbk.eu}
\and
Patricia Martín-Rodilla \\
IEGPS-CSIC, Spanish National Research Council, Santiago de Compostela, Spain \\
\texttt{p.m.rodilla@iegps.csic.es} }
\date{}

\begin{document}
\maketitle

\begin{abstract}
Argument Mining (AM) transforms natural language into its underlying argument structures. This transformation is typically realized through a sequence of AM tasks that form an end-to-end AM pipeline. However, AM approaches often differ in how they conceptualize these tasks, making direct comparisons between them difficult and opaque. This calls for a more nuanced, task-level analysis of AM approaches to enable clearer comparison and assessment.

This work presents a preliminary meta-study that systematically reviews several state-of-the-art end-to-end AM works and analyzes their pipelines through a triple-perspective framework---a linguistic, computational and domain perspective---to understand how the pipelines model arguments as structures, computes them, and integrates domain knowledge. We further propose a general design to the linguistic and computational perspectives, illustrating how key AM tasks are designed for modeling and computation of argument structures. Our proposed framework lays the groundwork for methodology-centered descriptions across AM approaches, facilitating deeper understanding and more systematic comparisons in future research.
\end{abstract}

\section{Introduction}

Argument Mining (AM) is a sub-field of computational argumentation concerned with the automated transformation of natural language into structured argument representations \citep{aroraArgumentMiningCategorical2023, lopescardosoArgumentationModelsTheir2023}. Early work in AM typically concentrated on individual tasks---such as detecting argument units \citep{trautmannFineGrainedArgumentUnit2020, luginiContextualArgumentComponent2020} or classifying argumentative relations \cite{joClassifyingArgumentativeRelations2021}---reflecting the high costs of annotation and data collection in argumentation. While such single-task approaches have achieved strong performance \cite{aroraArgumentMiningCategorical2023}, they provide only a narrow and fragmented view of argumentative discourse. Recent advances in resource-efficient natural language processing (NLP) methods, coupled with the availability of large-scale corpora \citep{lindahl_2020}, have shifted attention toward end-to-end AM approaches that cover the entire transformation process from raw discourse to complete argument structures \cite{ruiz-dolz.etal_2021}.

This shift has also enabled AM to expand beyond single-purpose tasks and into a wide range of applications and research domains. Improvements in long-context reasoning and discourse modeling now make it feasible to generate argument structures from longer and more diverse texts. This has resulted in increased integration of AM not only in downstream NLP tasks---such as opinion mining \citep{dragoniCombiningArgumentationAspectbased2018, alhindiFactVsOpinion2020}, stance classification \cite{aldayelStanceDetectionSocial2021}, fact-checking \cite{visserSkepticWebService2022}, and quality assessment \citep{garcia-gorrostietaCorpusArgumentAnalysis2019, kashefiArgumentDetectionStudent2023}---but also across a variety of research domains, including law \citep{alzubaerPerformanceAnalysisLarge2023, santinArgumentationStructurePrediction2023}, political science and sociology \cite{goffredoDISPUTool20Modular2023}, bio-science \citep{siBiomedicalArgumentMining2022, molinetAssessingArgumentbasedNatural2025}, and discourse analysis \cite{iraniArguSenseArgumentCentricAnalysis2024}. In parallel, practical applications of AM are also expanding, particularly in areas that require logical reasoning \cite{kashefiArgumentDetectionStudent2023}, dialogical interaction \cite{iraniArguSenseArgumentCentricAnalysis2024}, and decision-making \cite{brunAnalysingPracticalArgumentation2016}. The increased attention on AM motivates closer investigation into how AM approaches are generally conceived and practiced across domains and applications.

One common way to represent end-to-end AM is through high-level task representations that show how the transformation is operationalized. These representations, often referred to as \textbf{argument mining pipelines} \cite{stabIdentifyingArgumentativeDiscourse2014, lenzArgumentMiningPipeline2020, zhengKNOWCOMPPOKEMONTeam2024}, outline the tasks involved and their arrangement, providing a system-level view of how discourse is transformed into structured arguments. Yet pipelines have typically been presented only at the level of listing tasks, without deeper examination of how those tasks are ordered, how they interact, or what role they play in the conceptualization of the AM process itself. Furthermore, the tasks themselves are not standardized, differing in name, functionality, and application. This highlights the need for a more systematic review of the pipelines---one that investigates not only which tasks are included, but also how they are organized and what role they serve in the AM pipeline.

For this effort, we introduce the following \textbf{working definition}:

\begin{quote}
``An argument mining pipeline is an executable realization of an end-to-end argument mining that illustrates what tasks are required to transform raw discourse into a well-defined argument structure.”
\end{quote} 

Analyzing Argument Mining (AM) pipelines is challenging not only because of the complexity of the subtasks required to construct argument structures, but also due to the inherently multidisciplinary nature of the field. AM operates at the intersection of three principal disciplines: argumentation theory, computational methods, and domain knowledge. Argumentation theory formulates the identification, extraction, and structuring of argumentative content in discourse, while computational methods enable the automation of these processes at scale. Domain knowledge, in turn, contextualizes argument structures and supports knowledge discovery from argumentative interactions. Building on prior work that conceptualizes AM as sequences of tasks \cite{Baroni2018_ch12}, we propose to reinterpret AM pipelines through a complementary, non-sequential lens organized around three core objectives: modeling, computation, and domain integration. Accordingly, our study analyzes AM pipelines using a triple-perspective approach:

\begin{itemize}
    \item the \emph{linguistic perspective}, how pipelines conceptualize and represent argument structures;
    \item the \emph{computational perspective}, how pipelines automate and compute argument structures; and
    \item the \emph{domain perspective}, how pipelines integrate domain knowledge into its design and application.
\end{itemize}

Using this framework, we conduct a systematic review of representative end-to-end AM approaches, identifying core design choices and patterns in how argument structures are modeled, computed, and contextualized. Our goal is to provide a macro-level analytical basis for understanding and comparing contemporary AM pipelines.

\section{Related Work}
\label{sec:rel_work}

Existing meta-level surveys on Argument Mining (AM) have primarily organized the field around datasets \citep{cabrioFiveYearsArgument2018, lytosEvolutionArgumentationMining2019}, task taxonomies \citep{lawrenceArgumentMiningSurvey2020}, or application domains \citep{vecchiArgumentMiningSocial2021}. While these works have been instrumental in mapping the development of AM, they typically examine individual dimensions of the field rather than the design of complete pipelines.

In practice, AM operates as a structured process that transforms raw discourse into argument representations composed of \textbf{argument units} (e.g., claims, premises) connected by \textbf{argument relations} (e.g., support, attack). This transformation is realized through AM pipelines, whose design choices---such as how tasks are defined, ordered, and integrated---directly affect system behavior, interpretability and applicability. Despite this central role, there has been limited standardized macro-level analysis of how pipelines are designed or how their methodological choices can be systematically compared. As a result, the conceptual, computational, and contextual dimensions of AM systems are often evaluated in isolation.

Earlier efforts have partially acknowledged this. For example, \citet{Baroni2018_ch12} distinguished between linguistic modeling and computational realization of argument structures, framing AM as a process of resource construction followed by automation. However, pipeline designs have become increasingly diverse, expanding both the linguistic modeling and the computational landscape. A shift from predominantly theory-driven modeling toward more data-driven approaches is evident, accompanied by rapid advances in neural and language modeling techniques. Joint neural architectures now integrate unit detection and relation classification within shared representations \citep{eger.etal_2017, niculae.etal_2017}; knowledge-enriched systems incorporate discourse cues and domain ontologies \citep{al-khatibEndtoEndArgumentationKnowledge2020, jiIncorporatingDomainKnowledge2023}; and recent LLM-based methods collapse traditional modular pipelines into unified prompting frameworks \citep{gretz.etal_2020, joLLMArgument2023}. This heterogeneity further complicates systematic comparison and highlights the absence of a unified analytical lens.

In this work, we address this gap by proposing a macro-level, triple-perspective framework that systematically examines how end-to-end AM pipelines model, compute, and contextualize argument structures across domains.

\section{Selection of Relevant Literature}
\label{sec:rel_lit}

To conduct our systematic review of end-to-end AM pipelines, we retrieved potentially relevant literature from two major research databases: SCOPUS\footnote{\url{https://www.scopus.com/}} and Web of Science (WoS)\footnote{\url{https://www.webofscience.com/}}. Because the term pipeline is not consistently used in AM, our search queries also included related terms such as argument unit, extraction, detection, relation, support, and attack. The initial search returned 273 records from SCOPUS and 237 from Web of Science. After merging and removing duplicates based on title and abstract, we obtained 384 documents, referred to as the Argument Mining Pipelines original set (AMP384).

Next we defined the following selection criteria to identify the relevant literature:

\begin{enumerate}
    \item The document \textbf{must} explicitly discuss the AM process with intent to produce argument structures, whole or part.
    \item The tasks involved in transformation process \textbf{must} be identifiable.
    \item Its input data and output argument structure \textbf{must} be identifiable.
    \item The document should preferably be open-access for detailed analysis of its methodology, data and evaluation strategy.
\end{enumerate}

We employed GPT-4 to assist with abstract-level screening by providing a structured prompt with a set of questions based on the predefined selection criteria and the triple-perspective framework, described in the next section, and instructing it to return “unknown” where information was not explicitly stated. Based on its responses, we shortlisted 164 studies that most closely satisfied the criteria, forming the \textit{Argument Mining Pipelines 164 dataset} (AMP164). Inclusion in AMP164 does not imply that every study fully meets all selection criteria; rather, the dataset comprises works that positively answered most questions. The complete list of documents and additional documentation of the selection process are available on our public repository\footnote{\url{https://github.com/The-obsrvr/ArgumentMiningPipelines}}.

\begin{figure}
    \centering
    \includegraphics[width=0.7\textwidth]{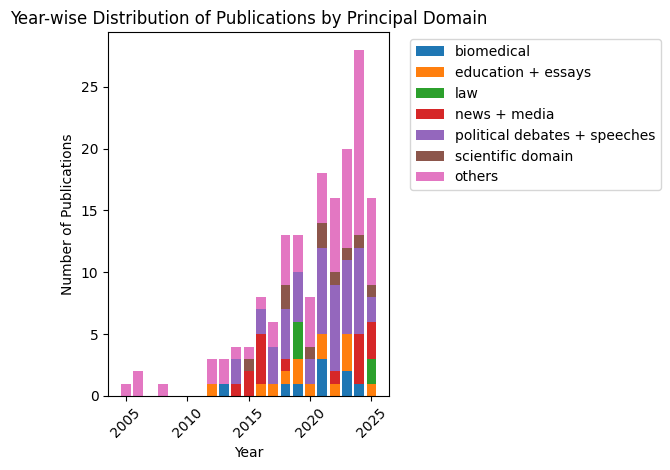}
    \caption{Distribution of the documents in the AMP164 underscoring the growth in AM pipelines over the last two decades (up to June 2025), further split against the principal domain in which the work lies.}
    \label{fig:pub_dist}
\end{figure}

Figure~\ref{fig:pub_dist} shows the yearly distribution of AMP164, indicating a rising shift toward end-to-end AM research. It also summarizes the principal application domains identified in our analysis: six domains were explicitly defined based on thematic grouping, while remaining studies were categorized under a default ``others" label when domain information was unclear. The distribution indicates a strong interest in domains characterized by multi-stance and deliberative discourse, including debates, essays, and social media.

In the following section, we present the main findings derived from analyzing AMP164 manually through our triple-perspective framework.

\section{Triple-perspective Framework}
\label{sec: trip_pers}

To apply our triple-perspective framework to the selected literature, we analyzed the principal design choices underlying AM pipelines from each respective lens. The linguistic and computational perspectives are examined through an investigative review, with findings and generalized representations of their core tasks and design choices presented in Sections~\ref{sec:ling_pers} and~\ref{sec:comp_pers}. The analysis is then extended to the domain perspective in Section~\ref{sec:dom_pers}, where we examine how domain knowledge influences pipeline design and operational processes.

\subsection{Linguistic Perspective}
\label{sec:ling_pers}

Linguistic perspective focuses on how the argument structure is conceptualized and modeled by the AM pipeline. The structure modeling can be seen as the product of conceptualizing, framing, and applying principles from argumentation theory in conjunction with linguistic principles to generate formal representations of argument structures. For a systematic evaluation of the linguistic perspective, we defined a set of guiding questions and applied them to our selected literature:

\begin{enumerate}
    \item What are the core argumentation theory and/or linguistic principles used in the structure modeling?
    \item How has the argument and its structure been formalized?
    \item How has the data resource, if any, been produced?
\end{enumerate}

Our review indicates that structure modeling is typically realized through two primary methodologies: \textit{theory-driven} and \textit{data-driven}.

Theory-driven approaches adopt established argumentation frameworks, most prominently Toulmin-inspired models \cite{karbach_1987} and Walton’s taxonomy of argument schemes \cite{waltonClassificationSystemArgumentation2015}. These frameworks provide predefined categories and relational structures that aid in the identification of argumentative units and their interactions \cite{morio.fujita_2018}. In contrast, data-driven approaches derive argument structures using indicators from the discourse, such as argumentative discourse markers \cite{oepen.etal_2016, lawrence.etal_2017}, context \cite{rocha.cardoso_2022}, or interactional patterns \cite{mestre.etal_2021}. Rather than strictly adhering to predefined theoretical taxonomies, these methods infer structure from linguistic signals and contextual prompts present in the data. While offering greater flexibility and adaptability to real-world discourse, data-driven methods may introduce interpretative variability due to the absence of a fixed theoretical foundation. Modern works are combining the two approaches in a hybrid methodology, where discourse markers are used to identify empirically arguments in the discourse which are then justified using existing theoretical foundations. 

Beyond methodological orientation, modeling decisions can also be examined across two analytical levels: the \textit{micro-level} and the \textit{macro-level}.

At the micro-level, pipelines identify and segment argumentative units within discourse, distinguishing them from non-argumentative segments. Units vary in granularity, ranging from token- or span-level representations \cite{kantesaria.p._2018, su.etal_2023} to sentence-level classifications \cite{hua.wang_2017}. They may further be categorized through argument component classification \cite{cao_2023}.

At the macro-level, modeling concerns the relational structure connecting argumentative units. Pipelines define and classify relations such as support, attack, rephrasing \cite{ruiz-dolz.etal_2025a}, agreement or disagreement with respect to a proposition  \cite{joClassifyingArgumentativeRelations2021}, or broader dialogical functions such as questioning, justification, acknowledgment, or summarization \cite{morio.fujita_2018, feltonCapturingDeliberativeArgument2022, vaitla.etal_2024}.

\begin{figure}[t]
    \centering
    \includegraphics[width=\linewidth]{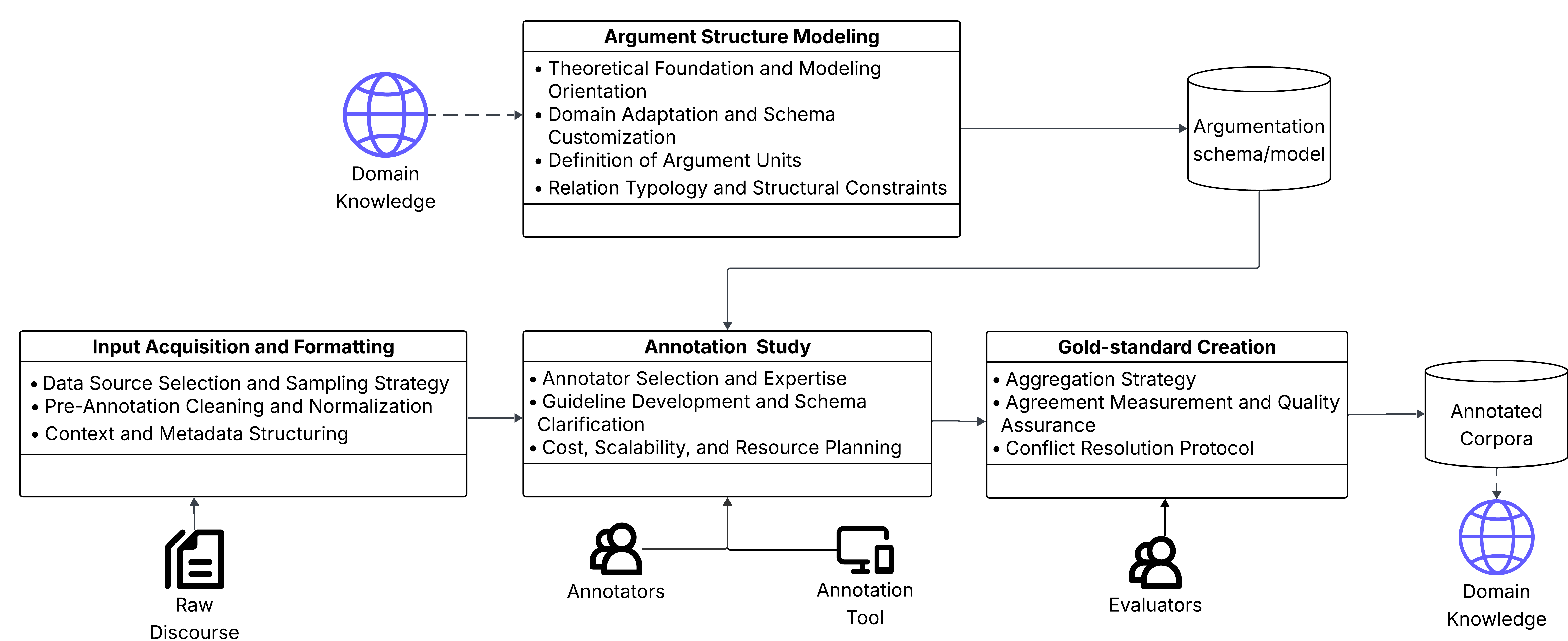}
    \caption{Overview of the key design choices made within the linguistic modeling component of the pipeline.}
    \label{fig:ling_stage_frame}
\end{figure}

The modeling choices are ultimately formalized in concrete resources---such as argumentation grammars and annotated corpora \cite{lindahlAnnotationComputationalArgumentation2024}---which shape how argumentative content is identified, structured, and replicated by the pipeline. Annotated corpora, in particular, play a central role: they both operationalize theoretical schemes in practice and provide reusable resources for downstream tasks such as pipeline automation and discourse analysis.

Corpus creation initiates with selecting appropriate data sources and pre-processing them according to predefined assumptions and constraints, often inferred from domain. The modeling framework is then applied through a structured annotation study, which specifies annotator roles, annotation platforms, and detailed guidelines \cite{visserAnnotatingArgumentSchemes2021}. Annotation quality is assessed using inter-annotator agreement metrics, after which annotations are aggregated---commonly through majority voting---to establish a gold standard. Disagreements or inconclusive cases are resolved through revision procedures \cite{musi.etal_2016}. This remains one of the most labor-intensive components of AM pipelines. (see also \cite{musi.etal_2016, lindahlAnnotationComputationalArgumentation2024, lopescardosoArgumentationModelsTheir2023}).

Figure~\ref{fig:ling_stage_frame} summarizes the general modeling processes and their key design choices inferred from this perspective. Examining structure modeling in this way potentially clarifies the theoretical assumptions made in the AM pipelines and highlights the associated resource costs in terms of data and human labor. It also reveals how modeling choices are often domain-sensitive, with annotation schemes and corpora differing substantially across legal, political, and other domains. In the following sections, we examine how these modeling decisions interact with computational and domain considerations.

\subsection{Computational Perspective}
\label{sec:comp_pers}

The computational perspective focuses on how argument structures are operationalized and automated at scale through computational methods. To study this systematically, we formulated the following guiding questions on our selected literature:

\begin{enumerate}
\item What are the main AM tasks automated in the work?
\item How are the identified AM tasks implemented computationally?
\item For each computational model of a task, what are their input and output requirements?
\item How has each identified computational model been developed and evaluated?
\end{enumerate}

Our review established two core design choices in the literature: (i) the modeling paradigm employed for automation, and (ii) the architecture and decomposition of tasks within the pipeline.

\textbf{Modeling paradigms.}
Argument structures are computed through three broad paradigms:
(i) \emph{feature-based approaches}, including discourse marker identification and manual feature engineering;
(ii) \emph{deep learning models}, particularly neural architectures and transformer-based systems; and
(iii) \emph{large language models (LLMs)} adapted to argumentation tasks, often in generative or instruction-tuned settings.

Feature-based approaches \cite{du.etal_2017, huwaidah.etal_2021, segura-tinoco.cantador_2023a} rely on explicit linguistic, syntactic, and discourse-level features derived from the underlying modeling framework and data. These approaches offer interpretability and close alignment with theoretical foundations but are limited in their scope and generalizability.

Deep learning approaches \cite{stylianou.vlahavas_2021, mayer.etal_2021} reduce dependence on handcrafted features by learning distributed representations directly from annotated data. Fine-tuned transformer models have become dominant in tasks such as argument component identification and relation classification, offering improved generalization and scalability.

More recently, LLM-based approaches \cite{cabessa.etal_2025a, clayton.etal_2024, otiefy.alhamzeh_2024} adapt AM tasks as generative or instruction-following problems, producing structured argument representations directly from raw discourse. While these models demonstrate flexibility and cross-task transfer, they introduce challenges related to output controllability, evaluation, interpretability, and computational cost.

Across the literature, there is a clear shift from feature-engineered approaches toward increasingly data-intensive neural approaches, driven by improvements in computational resources and data availability.

\textbf{Task architecture and decomposition}.
Pipelines differ in how argument structure prediction is decomposed into computational steps. To facilitate systematic comparison, we conceptualize each computational component as an \textbf{input–output (I/O) unit}: a modular element that receives structured input, processes it using a defined model, and produces a specified output. This abstraction allows heterogeneous pipelines to be analyzed within a unified framework.

\begin{figure}[t]
    \centering
    \includegraphics[width=\linewidth]{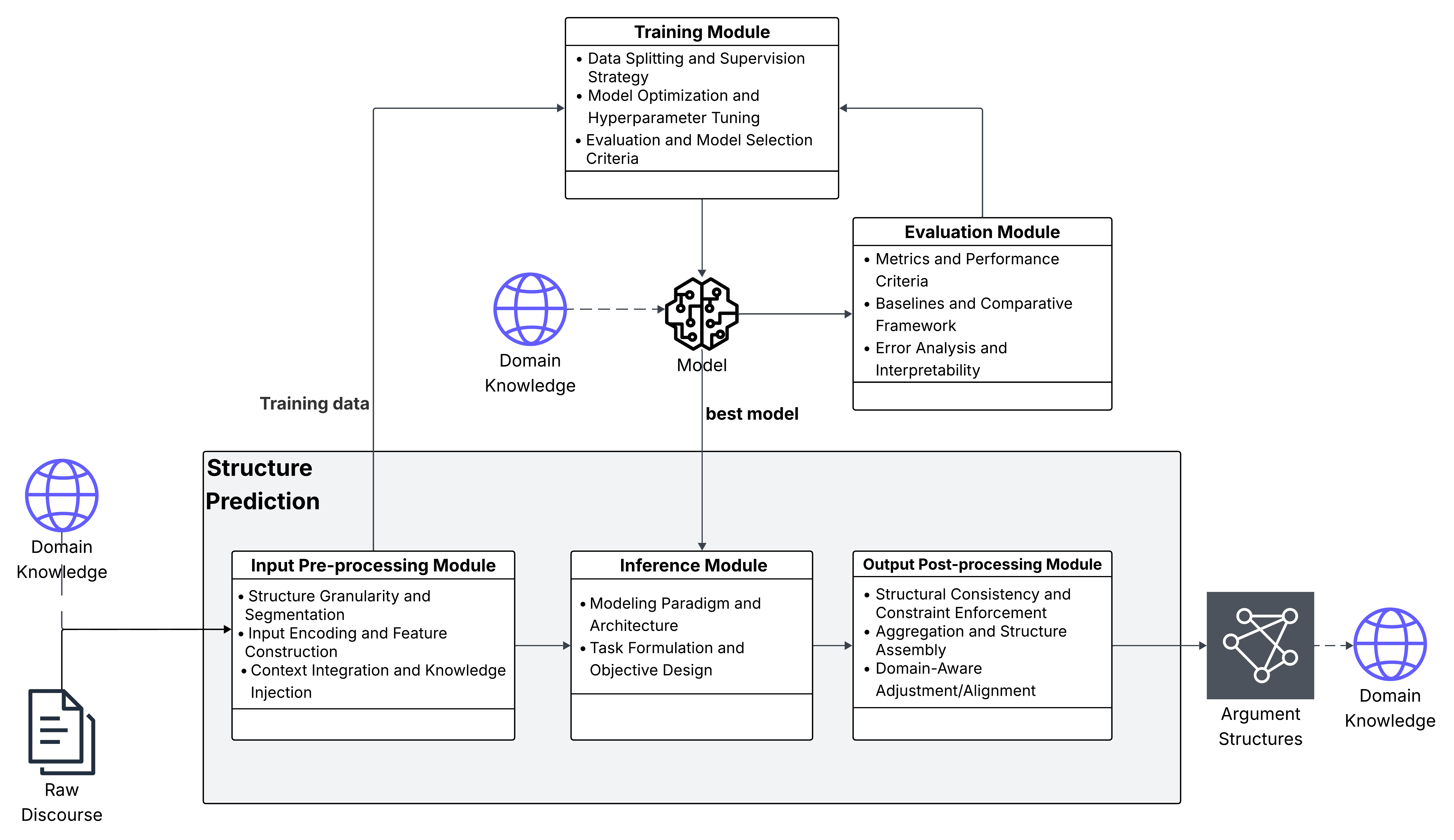}
    \caption{Overview of the key design choices made within the computational component of the pipeline.}
    \label{fig:comp_stage_frame}
\end{figure}

As illustrated in Figure~\ref{fig:comp_stage_frame}, an I/O unit typically begins with input formatting, determined by (i) the AM task (e.g., extraction, classification, relation identification), (ii) the granularity of representation, and (iii) the learning strategy (the modeling paradigm). The formatted input is processed by the task-specific model, which optionally may require training and optimization. Predictions may then undergo post-processing to integrate intermediate outputs into a coherent argument structure.

Based on their decomposition strategy, AM pipelines generally adopt either a \textbf{multi-step design} or a \textbf{single-step `unified' design}. Multi-step architectures generate intermediate representations through sequential I/O units \cite{stylianou.vlahavas_2021, su.etal_2023}, enabling modular evaluation and interpretability. However, they require additional engineering effort and task-specific supervision. In contrast, single-step architectures aim to produce complete argument structures directly from raw input \cite{du.etal_2017, sun.etal_2024a}. While potentially less transparent, these approaches exploit cross-task dependencies and require relatively lesser computational resources.

Together, these two choices---modeling paradigm and task decomposition---capture the principal computational design choices observed across the AM pipelines.


\subsection{Domain Perspective}
\label{sec:dom_pers}

The linguistic and computational perspectives highlight how argument structures are modeled and automated. However, a comprehensive assessment of AM pipelines also requires examining how they engage with domain knowledge \cite{lauscherScientiaPotentiaEst2022}. Domain influences pipelines in two complementary ways: it shapes modeling and computational choices, and it determines how extracted argument structures are interpreted or applied.

In this context, domain refers to the discourse setting and knowledge structures underlying the data—its genre, conventions, terminology, and ontological assumptions—which influence how arguments are expressed and understood \cite{lauscherScientiaPotentiaEst2022}. Domains may be limited in scope (e.g., classroom discussions or single conversations) or broad and discipline-level (e.g., legal, financial, political, biomedical). Each introduces specific knowledge requirements ranging from commonsense assumptions to highly specialized expertise.

The domain perspective examines both the forms of knowledge integrated into AM pipelines and the mechanisms through which they influence modeling assumptions, computational strategies, and evaluation criteria. As reflected in Figure~\ref{fig:pub_dist}, the AMP164 data spans diverse domains, particularly those characterized by deliberation and multi-stance discourse, such as debates, politics, social media, and healthcare.

\textbf{(1) Schema Design.}
Domain knowledge directly constrains how argument units and relations are defined. Terminology, ontologies, and discourse conventions determine what qualifies as a claim or premise and which relations---such as support or attack---are meaningful. For example, ontology-guided approaches have been applied in biomedical, financial and legal contexts \cite{siBiomedicalArgumentMining2022, alzubaerPerformanceAnalysisLarge2023, liuAntCriticArgumentMining2024}, while broader commonsense or factual knowledge can shape relation classification \cite{botschenFrameEntityBasedKnowledge2018, joClassifyingArgumentativeRelations2021}. As a result, annotation schemas and modeling assumptions often vary substantially across domains.

\textbf{(2) Computational Modeling and Application.}
Domain knowledge also guides how argument structures are computed and subsequently used. Structured representations can support downstream tasks such as stance analysis \cite{ruckdeschelArgumentMiningAttack2024}, implicit reasoning detection \cite{schaeferImprovingImplicitStance2019, stahlMindGapAutomated2023}, debate outcome prediction  \cite{hsiaoModelingInterAttack2022}, and fallacy identification \cite{macagnoArgumentationSchemesFallacies2022, ruiz-dolzDetectingArgumentativeFallacies2023}. In such cases, argument structures function not only as intermediate representations but also as analytical tools for domain-specific knowledge discovery.

\textbf{(3) Evaluation and Post-processing Strategies.}
Finally, domain also affects the evaluation practices adopted in the pipeline. Performance must be assessed not only by predictive accuracy but also by how well extracted structures capture domain-specific reasoning norms and patterns. Specialized domains such as law, finance, or biomedicine require evaluation criteria aligned with domain standards \cite{groza.popa_2016, alzubaerPerformanceAnalysisLarge2023, sperrle.etal_2019, liuAntCriticArgumentMining2024}. In contrast, political and social domains may require sensitivity to stance diversity, implicit reasoning, and potentially offensive or sensitive content. Domain knowledge could further guide post-processing decisions, such as enforcing structural constraints, filtering implausible relations, or validating outputs against ontologies and expert rules.

Overall, incorporating the domain perspective clarifies how AM pipelines are situated within specific discourse contexts and highlights the conditions under which modeling and computational choices remain valid and accurate. A more systematic treatment of domain interaction remains an important direction for future research, particularly for improving transferability of argument resources across contexts.

\section{Significance of Pipeline Comparison}
\label{sec:assessment}

The previous sections examined AM pipelines through a multi-perspective lens, revealing substantial variation in pipeline design choices. Considering these dimensions collectively offers a more systematic basis for comparing pipeline designs. Rather than viewing pipelines solely as modular sequences of tasks, the framework encourages analysis of how linguistic, computational, and domain-related design choices interact within a system. This framework supports three analytical benefits:

\textbf{(1) Pipeline-level characterization.}  
Pipelines can be grouped according to shared methodological features---such as on their level of task decomposition, modeling paradigm adopted, structure modeling choices---enabling better benchmarking and resource-sharing between them.

\textbf{(2) Clarification of methodological dependencies.}  
The framework highlights how decisions in modeling, computation, and domain integration influence one another. Making these dependencies explicit supports more transparent interpretation of pipeline behavior and its reproduction.

\textbf{(3) Systematic comparison of pipelines.}  
By assessing the pipelines through the same analytical framework, we enables comparison beyond performance metrics or architectural choices. Approaches with similar results may differ in their theoretical grounding, task decomposition, or domain integration.

Overall, a holistic assessment visualizes AM pipelines not as sequences of independent tasks but as dynamic, methodological configurations whose linguistic, computational, and domain characteristics jointly determine their behavior, scope and application. Such a perspective supports clearer comparison and contributes to a more structured understanding of the evolving landscape of end-to-end AM systems.

\section{Conclusion}

This study examined AM through the lens of end-to-end pipelines, focusing on how argument structures are modeled, computed, and contextualized within a domain. Using a triple-perspective framework---linguistic, computational, and domain---we conducted a structured review of representative AM systems to identify core design choices and analyzed how these choices shape pipeline behavior.

Our findings indicate a shift toward more structured and hybrid configurations that move beyond isolated subtasks. Modeling assumptions, computational strategies, and domain requirements emerge as interdependent factors that jointly shape pipeline behavior and applicability. Recognizing these interactions is essential for developing adaptable and methodologically transparent AM systems.

While this work provides a macro-level perspective on pipeline design, it is limited in its depth of analysis. Each perspective and its design choices require deeper investigation. Future research should further refine and align modeling, computation, and domain integration to support resource reuse, cross-domain transfer, and more coherent evaluation. Advancing in this direction will help build AM pipelines that are robust, reproducible, and better suited to real-world discourse analysis.

\section*{Acknowledgements}
This research work has received funding from the European Union's Horizon Europe research and innovation programme under the Marie Skłodowska-Curie Grant Agreement No. 101073351. Views and opinions expressed are however those of the author(s) only and do not necessarily reflect those of the European Union or European Research Executive Agency (REA). Neither the European Union nor the granting authority can be held responsible for them.

{\small
\bibliographystyle{plainnat}
\bibliography{sample}         
}

\end{document}